\documentclass[10pt, a4paper, twocolumn, showabstract]{naverlabseurope}

\usepackage{lipsum}
\usepackage{multicol}
\usepackage{tikz,tkz-kiviat,pgfplots}

\usepackage{multirow}
\usepackage[table,xcdraw]{xcolor}
\usepackage[ruled,vlined]{algorithm2e}
\usepackage{CJKutf8}
\usepackage{arydshln}

\newcommand{\fix}[1]{\textcolor{red}{ #1}}

\newcommand{\seamless}{\texttt{SeamlessM4T-v2-large}}
\newcommand{\llama}{Llama-3.1-8B-Instruct}
\newcommand{\eurollm}{EuroLLM-9B-Instruct}
\newcommand{\speechmapper}{SpeechMapper}

\newcommand{\ours}{SpeechMapper: Speech-to-text Embedding Projector for LLMs}

\title{SpeechMapper: Speech-to-text Embedding Projector for LLMs}
\titlerunning{\ours}

\correspondingauthor{marcely.zanon-boito@naverlabs.com}

\authors{Biswesh Mohapatra$^{1\star}$ \authsep Marcely Zanon Boito$^{2\star}$ \authsep Ioan Calapodescu$^{2}$}
\affiliations{$^{1}$Inria Paris, France \authsep $^{2}$NAVER LABS Europe}
\contributions{$^{\star}$equal contribution}

\begin{abstract}
Current speech LLMs bridge speech foundation models to LLMs using projection layers, training all of these components on speech instruction data. This strategy is computationally intensive and susceptible to task and prompt overfitting. 
We present \textit{SpeechMapper}, a cost-efficient speech-to-LLM-embedding training approach that mitigates overfitting, enabling more robust and generalizable models.
Our model is first pretrained without the LLM on inexpensive hardware, and then efficiently attached to the target LLM via a brief 1K-step instruction tuning~(IT) stage. 
Through experiments on speech translation and spoken question answering, we demonstrate the versatility of SpeechMapper's pretrained block, presenting results for both \textit{task-agnostic} IT, an ASR-based adaptation strategy that does not train in the target task, and task-specific IT.
In task-agnostic settings, Speechmapper rivals the best instruction-following speech LLM from IWSLT25, despite never being trained on these tasks, while in task-specific settings, it outperforms this model across many datasets, despite requiring less data and compute. Overall, SpeechMapper offers a practical and scalable approach for efficient, generalizable speech-LLM integration without large-scale IT.
\end{abstract}

\begin{document}

\maketitle

\section{Introduction}

LLMs have demonstrated unprecedented capabilities on text-based tasks~\cite{gpt4, llama, jiang2024mixtral}, motivating research into extending them to other modalities~\cite{team2023gemini,defossez2024moshi,hu2024wavllm, huang2024audiogpt}. For speech, state-of-the-art speech foundation models~(SFM) ~\cite{whisper,barrault2023seamlessm4t} excel at automatic speech recognition (ASR) and speech translation (ST) by leveraging large text decoders that function as internal LMs, but broader spoken language understanding tasks, such as spoken question answering~(SQA), require adaptation. To address this gap, recent works have explored integrating SFMs with LLMs. The simplest approach cascades ASR with an LLM, which is effective but discards acoustic cues and adds latency. Similarly, mapping speech into discrete tokens reduces the modality gap but is slow and sensitive to quantization quality~\cite{rubenstein2023audiopalm,nguyen2024spirit,ambilduke2025}, underscoring the need for more efficient, content-preserving speech-to-LLM integration. An alternative line of research focuses on directly bridging the SFM encoder outputs to LLMs: early methods inject speech via cross-modal adapters or prepend speech embeddings to the token sequence~\cite{whispering-llama, Fathullah2023PromptingLL}, whereas complex systems jointly instruction-tune~(IT) both the SFM encoder and the LLM across multiple tasks~\cite{hu2024wavllm, salmonn, tan2024ssr}. Despite their effectiveness, these methods are resource-intensive, require high-end GPUs and large speech datasets, and the added adapters can lead to prompt overfitting, reducing generalization to unseen tasks.

In contrast to joint SFM–LLM training, recent work maps speech into embeddings that the LLM can consume directly, preserving its original text capabilities. Wav2Prompt~\cite{wav2Prompt} trains a speech encoder to generate LLM-like embeddings, enabling zero-shot generalization but still requiring full LLM forward passes during training. SSR~\cite{tan2024ssr} similarly learns a speech-to-LLM projector, but through costly distillation with an explicit aligner, followed by LLM fine-tuning.
The proposed \textit{SpeechMapper} follows this line of work. We remove key limitations by avoiding sequence aligners and reducing reliance on costly LLM forward passes, while still producing a projector that generalizes to unseen speech-to-text tasks.

Our SpeechMapper projector maps SFM embeddings directly into the target model’s embedding space through a two-stage training procedure: pretraining and adaptation. Pretraining uses only the LLM's embedding layer, decoupling training from its costly forward computations, thereby scaling independently of LLM size and enabling the use of more affordable V100 GPUs.
Adaptation then rapidly fine-tunes a pretrained SpeechMapper in either a \textit{task-agnostic} or \textit{task-specific} manner, keeping the core LLM unchanged. This produces, respectively, a zero-shot speech LLM or a task-specific specialist model, all in just 1.5\,h of training on an A100 GPU for an 8–9B LLM. Across ST and SQA experiments, SpeechMapper matches or surpasses the best instruction-following speech LLM from the IWSLT25 short track~\cite{agostinelli-etal-2025-findings}, while requiring far less data and compute, offering a highly resource-efficient path to speech-enabled LLMs.
\begin{figure*}[h]
  \centering
  \includegraphics[width=\linewidth]{
  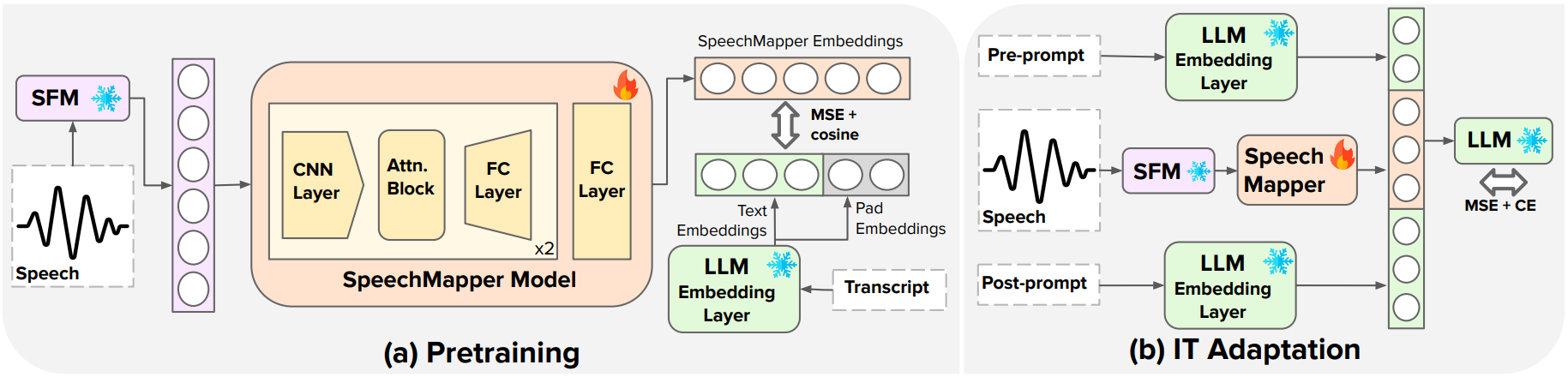}
\caption{\speechmapper{} two-stage training. \textbf{(a) Stage 1: Pretraining} maps SFM embeddings to shorter sequences trained to match the corresponding LLM text embeddings. Pad embeddings ensure equal sequence lengths. \textbf{(b) Stage 2: IT Adaptation} connects the pretrained projector to the LLM, enabling zero-shot speech capabilities after just 1K steps.}
  \label{fig:Architecture}
\end{figure*}
\section{SpeechMapper}\label{sec:methodology}

SpeechMapper training proceeds in two stages: pretraining with the frozen LLM embedding layer, followed by rapid adaptation with the full LLM kept frozen. Our experiments will show that pretraining alone yields a competent zero-shot speech-to-LLM projector, indicating most of the mapping can be learned without costly LLM forward passes. Adaptation then refines the embeddings, boosting performance while avoiding task- or prompt-specific overfitting and leaving the LLM unchanged. We now present SpeechMapper's architecture and training procedure.

\subsection{Architecture}
Fig.~\ref{fig:Architecture}~(a) shows the SpeechMapper architecture. Input speech is first encoded by a frozen SFM, and the resulting embeddings serve as input to the projector.\footnote{Since the SFM is not updated during training, embeddings can be precomputed, reducing memory and computation during training.} Speech embeddings have lower dimensionality than LLM embeddings, but the sequences are considerably longer. To handle this, the projector consists of two sequential blocks that progressively compress the input sequence, apply attention mechanisms, and map the representations into a higher-dimensional space. A final feed-forward layer produces the output used for loss computation. During inference, SpeechMapper generates speech embeddings that are directly injected into the LLM. More details regarding this architecture are given in Appendix Section~\ref{app:architecture}.

\subsection{Stage 1 --  Pretraining}
SpeechMapper is trained using ASR data only. Training minimizes the mean squared error~(MSE) loss between the projector’s output embeddings and the corresponding LLM embeddings of the target sentences. A direct MSE application is challenging due to the length mismatch: speech sequences are significantly longer than their text counterparts. Similar works~\cite{wav2Prompt, tan2024ssr} handle this via explicit alignment. Instead, we pad the target embedding sequence with a special pad token to match the projector’s output length. Our assumption is that by forcing some speech vectors to correspond to pad embeddings, the model is forced to shift semantic information toward the beginning of the sequence, implicitly modeling sequence length. Thus, during pretraining, SpeechMapper learns to produce the target embeddings followed by an arbitrary number of pad embeddings.\footnote{Successive compression in SpeechMapper keeps the ratio of target to pad embeddings close to or below 1:1.} The pad embedding is a reserved LLM token, allowing the LLM to ignore it during inference without additional processing. 

In preliminary experiments, we observed that once the first pad token is produced, the model tends to generate subsequent pads easily. To prevent undue emphasis on padding, we split the MSE loss into two parts: a higher-weighted $\text{MSE}_{\text{word}}$ for 
words and the first pad token, and a lower-weighted $\text{MSE}_{\text{pad}}$ for the remaining pads (Eq.~\ref{eq:mse}), with $\alpha\in{[1,9]}$ controlling the word-to-pad ratio.
Masking ensures each loss only considers the relevant embeddings. Following \cite{tan2024ssr}, we also incorporate the standard cosine embedding loss to emphasize critical embedding dimensions, which we found to be beneficial. Eq.~\ref{eq:final_loss} 
presents the stage~1 loss, where $\gamma$ controls the cosine weight.

\subsection{Stage 2 -- IT adaptation}
After stage~1 pretraining, we adapt the pretrained projector atop a frozen LLM (Fig.~\ref{fig:Architecture}~(b)). This stage runs for 1K steps using the loss defined in Eq.~\ref{eq:it}, which combines the cross-entropy~(CE) between the LLM's output and the target text with the previously defined $L_{\text{MSE}}$, computed between SpeechMapper's output and the text embeddings of the speech transcript.
The weighting parameter $\sigma$ controls the MSE contribution: a stronger MSE term prevents the projector from overfitting to the training prompts or tasks, while $\sigma=0$ reduces to standard LLM IT. By adjusting $\sigma$, we can build either task-specific models ($\sigma=0$), which tend to be overfit to the training task, or task-agnostic models ($\sigma>0.8$), which are capable of performing unseen tasks.

\begin{equation}\label{eq:mse}
    L_{\text{MSE}} = \alpha \, \text{MSE}_{\text{word}} + (10 - \alpha) \, \text{MSE}_{\text{pad}} 
\end{equation}
\begin{equation}\label{eq:final_loss}
    L_{\text{stage1}} = L_{\text{MSE}}- \gamma \, L_{\text{cosine}}
\end{equation}
\begin{equation}\label{eq:it}
    L_{\text{stage2}} =  (1 - \sigma)\,L_{\text{CE}} + \sigma\,L_{\text{MSE}}
\end{equation}
\section{Experimental Settings}\label{sec:experimentalsettings}

For our experiments, we consider three adaptation variants of stage~2: training on ASR data in either (i) a standard IT setting with ASR data~(ASR CE); (ii) adding $L_{MSE}$ to this IT stage~(ASR CE+MSE);  or (iii) training on ST/QA data in a task-specific IT setting~(ST/SQA CE). Comparing (i) and (ii) allows us to evaluate how the inclusion of MSE during task-agnostic adaptation facilitates zero-shot generalization,\footnote{Although CE-only IT setups typically yield task-specific systems, the very short adaptation step prevents \textit{severe} overfitting in practice.} while comparing them with (iii) highlights SpeechMapper's flexibility, demonstrating its ability to serve both as a pretrained block for versatile zero-shot speech LLMs, and as a foundation block for stronger task-specialized speech models. We now detail our experimental setting.

\noindent\textbf{Backbones:}
We use seamless-m4t-v2-large~\cite{barrault2023seamlessm4t} as the SFM,\footnote{We also experimented with speech-only pretrained models~\cite{NEURIPS2020_92d1e1eb,boito2024mhubert}, but stage~1 results were considerably worse. Results are presented in Appendix Section~\ref{section:apdx_sm_stage1}, Table~\ref{tab:lsonlyresults}.} extracting features from the 24th speech encoder layer, and averaging every two frames before feeding them to our projector.
We train SpeechMapper for two LLM backbones: \llama{}~\cite{llama} and \eurollm{}~\cite{martins2024eurollm}. Although stage~1 is free of LLM forward computation and can scale to larger models, we restrict our experiments to 8–9B LLMs because of the many required inferences.

\noindent\textbf{Data:}
SpeechMapper is pretrained and adapted in task-agnostic settings using the 960\,h LibriSpeech~(LS) corpus~\cite{panayotov2015librispeech}. For task-specific ST/SQA IT, we use the training set used in our baseline~\cite{lee-etal-2025-naver}, sampling a random subset since models train for only 1K steps~(considerably less than one epoch). 
For evaluation, ST experiments use EuroParlST~\cite{iranzo2020europarl} and CoVoST2~\cite{wang2020covost}, while SQA experiments use SpokenSQUAD~\cite{lee2018spoken} and LibriSQA~\cite{zhao2024librisqa}. Our prompts are presented in Appendix Section~\ref{sec:prompts}.

\noindent\textbf{SpeechMapper training settings:}
The model has 277M parameters. \textbf{Stage~1} training is implemented on the pasero library~\cite{pasero}, leveraging speech support from the IWSLT23 recipe~\cite{gow-smith-etal-2023-naver}. MMS normalization~\cite{pratap2024scaling} is applied to the target text before generating the text embeddings. Each SpeechMapper block~(Fig.~\ref{fig:Architecture}~(a)) consists of one convolutional layer (kernel size=6, stride=2), six Transformer layers, and a feed-forward layer (1024$\rightarrow$2048, 2048$\rightarrow$4096). We train using AdamW optimizer with cosine scheduler~(100K warm-up steps), dynamic batching, $\alpha$=~$5$, $\gamma$=~$100$ and a learning rate of $1e-4$ for 2M steps on 4$\times$V100-32GB~(4 days). 
\textbf{Stage 2} adaptation is implemented using the torchtune library~\cite{torchtune}. We load a trainable SpeechMapper atop a frozen LLM and perform IT with a constant learning rate scheduler, a learning rate of $1e-4$ or $5e-5$, gradient accumulation of 8, and batch size of 12. Models are trained on 1$\times$A100-80GB GPU for 1K steps, corresponding to approximately 1.5\,h of training time for an 8-9B LLM. Results are reported for \textit{CE} IT~($\sigma$~=~$0$, no MSE), and \textit{CE+MSE} IT~($\sigma$~=~$0.9$ and $\alpha$~=~$5$).\footnote{Appendix Sections~\ref{section:apdx_sm_stage1} and \ref{appendix:sec:stage2hyperparameters} discuss hyper-parameters for stage 1 and 2.}

\noindent\textbf{Baseline Comparison:} 
We compare SpeechMapper against the top-performing system from the IWSLT25 instruction-following~(IF) short track~\cite{agostinelli-etal-2025-findings}, where participants developed English speech LLMs for ASR, ST, and SQA. The winning submission~\cite{lee-etal-2025-naver}, denoted as \textbf{BEST-IWSLT25-IF}, integrates a transformer projector and LoRA adapters, trained on 2K hours of multitask and multimodal data~(5 days on 2$\times$A100-80GB GPUs, 58M parameters). This comparison is particularly meaningful for three reasons: (i) it uses the same SFM and LLM as SpeechMapper, (ii) its training results in a model highly specialized to the seen tasks and prompts, whereas our approach focuses on minimizing prompt and task overfitting, enabling greater flexibility to generalize to unseen formats, and (iii)~as the best system in the competition, it provides a strong and competitive baseline. Lastly, we do not compare with the similar Wav2Prompt~\cite{wav2Prompt} and SSR~\cite{tan2024ssr}, as they were not publicly available for inference.

\noindent\textbf{Inference and evaluation:} 
LLM inference for the toplines (transcripts/ASR+LLM), BEST-IWSLT25-IF, and our model are performed using the transformers library~\cite{wolf2019huggingface} in greedy mode with a maximum of 150 tokens.
We evaluate models on ST and SQA. For ST we compute COMET\footnote{Unbabel/wmt22-comet-da}~\cite{rei-etal-2022-comet} using both the reference transcription and translation, and by multiplying scores by 100 for increased readability. We believe COMET is a better metric than BLEU4 for evaluating LLM translations, since BLEU4 heavily penalizes rephrasing. For SQA, we follow the same LLM-as-judge protocol and scripts from BEST-IWSLT25-IF: evaluation is framed as a binary “yes/no” equivalence check using BERGEN~\cite{rau-etal-2024-bergen}. This setup presents the reference context, question, and both generated and reference answers. We report average accuracy across four LLM judges: \eurollm{}, Gemma3-12B/27B-Instruct~\cite{gemma3}, and Llama3.1-70B-Instruct. Presented results for stage~2 models correspond to averages obtained by running the IT stage twice using different seeds.

\section{Experiments and Results}\label{sec:experiments}

We now demonstrate the versatility of SpeechMapper across ST~(Sec.~\ref{experiments:st}) and SQA~(Sec.~\ref{experiments:sqa}) in zero-shot and in-domain settings. Despite pretraining only on LS and adapting for only 1K steps, our zero-shot and task-specific models rival BEST-IWSLT25-IF, a strong specialist speech LLM.

\subsection{Speech Translation}\label{experiments:st}

Table~\ref{tab:stnew} presents ST COMET results for EuroParl and CoVoST2 datasets. The top portion presents the toplines~(Transcripts+LLM), the middle portion our baselines (Seamless ST and BEST-IWSLT25-IF), and the bottom portion SpeechMapper models built with either EuroLLM or Llama. For each model, we report results for stage~1 pretraining and stage~2 adaptation under three settings: (i)~zero-shot ASR CE; (ii)~zero-shot ASR CE+MSE; and (iii)~task-specific ST~CE.  

We observe that the pretrained SpeechMapper (stage~1), regardless of the LLM backbone, already achieves strong ST performance, despite being trained on ASR data only. With EuroLLM, stage~1 outperforms the strong Seamless ST baseline for two EuroParl language pairs (en-fr, en-de), while for Llama-based models, stage~1 surpasses Seamless ST on en-de and en-it. Stage~2 further improves performance under task-agnostic and/or task-specific adaptation. We next discuss the differences between these adaptation approaches.

For \textbf{zero-shot ST} (i and ii), both CE and CE+MSE achieve comparable scores, while exhibiting strong ASR performance.\footnote{LS test-clean WER (CE/CE+MSE): 2.9/3.1 EuroLLM; 2.9/4.9 Llama. Results are presented in Appendix Section~\ref{appendix:otherresults}.} To better assess qualitative differences between the two objectives, we performed language identification using the langdetect toolkit on the outputs of the eight models~(two seeds per setting), reporting average accuracy for both the target language and English.
For Llama, CE shows limited target-language adherence (56.6\%), frequently defaulting to English (31.5\%), whereas CE+MSE markedly improves adherence to 87\% and reduces English output to 2.3\%. For EuroLLM, gains are more modest but consistent, with adherence increasing from 81.9\% to 85.2\% and English output decreasing from 6.6\% to 3.3\%.\footnote{Results are presented in Appendix Table~\ref{app:tab:languageconfusion}.} Overall, while ST scores are comparable, CE+MSE yields more reliable zero-shot behavior, reducing prompt violations and unintended ASR output.

The best zero-shot SpeechMapper~(EuroLLM + Stage~2 [ASR CE+MSE]) outperforms Seamless ST across three EuroParl language pairs, despite never being trained on ST data. Our \textbf{in-domain ST} further improves performance by an average of 6.9 COMET points, closing the gap with BEST-IWSLT25-IF and demonstrating the versatility of our task-agnostic pretrained block: by quick in-domain IT, we are able to build a model that rivals strong baselines trained for much longer time and far more data. Lastly, compared to Wav2Prompt's reported BLEU4 scores on EuroParl en-es and en-fr~(13.8/9.2)~\cite{wav2Prompt}, our approach already substantially outperforms it after stage~1 alone (27.8/24.8).\footnote{\scriptsize SacreBLEU (v2.3.1): nrefs:1$|$case:mixed$|$eff:no$|$tok:13a$|$smooth:exp}

\begin{table}
\resizebox{\columnwidth}{!}{\begin{tabular}{clcccc|cc}\toprule
&& \multicolumn{4}{c}{\textbf{EuroParl}}& \multicolumn{2}{c}{\textbf{CoVoST2}} \\
&& \multicolumn{1}{c}{\textbf{en-es}} & \multicolumn{1}{c}{\textbf{en-fr}} & \multicolumn{1}{c}{\textbf{en-de}} & \multicolumn{1}{c|}{\textbf{en-it}} & \textbf{en-de}& \textbf{en-zh}\\\midrule
& Transcripts + EuroLLM 9B \footnotesize{(topline)}& 85.9& 85.0& 82.5& 86.0& 78.3& 80.0\\
\multirow{2}{*}{}  & Transcripts + Llama 3.1 8B \footnotesize{(topline)}& 82.8& 81.0& 81.2& 84.1& 82.0& 77.0\\\midrule
\multirow{2}{*}{} & Seamless ST \footnotesize{(in-domain)}& 80.4& 74.8& 70.0& 76.0& \underline{83.0}& \underline{82.0}\\
& BEST-IWST25-IF \footnotesize{(in-domain)} & \underline{83.5}& \underline{81.1}& \underline{84.0}& \underline{86.0}& 78.9& 80.7\\\midrule
\multirow{4}{*}{\rotatebox[origin=c]{90}{\footnotesize{EuroLLM}}}   & Stage 1 \footnotesize{(zero-shot)}& 73.5& 76.0& 74.1& 75.8& 64.2        & 64.8        \\
& Stage 2 [ASR CE] \footnotesize{(zero-shot)}     & 78.6{\footnotesize\color{gray}$\pm0.5$}& 76.4{\footnotesize\color{gray}$\pm0.7$}& 72.1{\footnotesize\color{gray}$\pm2.0$}& 75.7{\footnotesize\color{gray}$\pm0.6$}& 70.2{\footnotesize\color{gray}$\pm0.9$}& \underline{74.0}{\footnotesize\color{gray}$\pm0.04$}\\
& Stage 2 [ASR CE+MSE] \footnotesize{(zero-shot)} & \underline{79.9}{\footnotesize\color{gray}$\pm1.1$}& \underline{77.4}{\footnotesize\color{gray}$\pm0.8$}& \underline{74.3}{\footnotesize\color{gray}$\pm2.1$}& \underline{78.4}{\footnotesize\color{gray}$\pm1.8$}& \underline{71.3}{\footnotesize\color{gray}$\pm0.7$}& 72.0{\footnotesize\color{gray}$\pm0.1$}\\
& Stage 2 [ST CE] \footnotesize{(in-domain)}& \textbf{85.4}{\footnotesize\color{gray}$\pm0.4$}& \textbf{84.5}{\footnotesize\color{gray}$\pm0.5$}& \textbf{82.2}{\footnotesize\color{gray}$\pm0.3$}& \textbf{85.5}{\footnotesize\color{gray}$\pm0.6$}& \textbf{77.0}{\footnotesize\color{gray}$\pm0.1$}&\textbf{79.9}{\footnotesize\color{gray}$\pm0.02$}        \\\arrayrulecolor{gray}\cdashline{2-8}\arrayrulecolor{black}
\multirow{4}{*}{\rotatebox[origin=c]{90}{\footnotesize{Llama 3.1}}} & Stage 1 \footnotesize{(zero-shot)}& \underline{76.4}& \underline{73.9}& \underline{72.3}& \underline{76.8}& \underline{67.1}& \underline{69.3}\\
& Stage 2 [ASR CE] \footnotesize{(zero-shot)}& 70.4{\footnotesize\color{gray}$\pm2.9$}& 69.4{\footnotesize\color{gray}$\pm2.2$}& 60.5{\footnotesize\color{gray}$\pm7.5$}& 63.9{\footnotesize\color{gray}$\pm6.8$}& 63.4{\footnotesize\color{gray}$\pm5.4$}& 62.2{\footnotesize\color{gray}$\pm8.2$}\\
& Stage 2 [ASR CE+MSE] \footnotesize{(zero-shot)} & 74.7{\footnotesize\color{gray}$\pm2.7$}& 71.0{\footnotesize\color{gray}$\pm2.8$}& 66.4{\footnotesize\color{gray}$\pm2.6$}& 73.2{\footnotesize\color{gray}$\pm2.6$}& 63.7{\footnotesize\color{gray}$\pm1.0$}& 68.6{\footnotesize\color{gray}$\pm1.5$}\\
& Stage 2 [ST CE] \footnotesize{(in-domain)}     & \textbf{84.5}{\footnotesize\color{gray}$\pm0.2$}& \textbf{82.4}{\footnotesize\color{gray}$\pm0.1$}& \textbf{80.9}{\footnotesize\color{gray}$\pm0.2$}& \textbf{84.5}{\footnotesize\color{gray}$\pm0.1$}& \textbf{75.5}{\footnotesize\color{gray}$\pm0.1$}& \textbf{78.6}{\footnotesize\color{gray}$\pm0.1$}\\\bottomrule
\end{tabular}}
\caption{ST COMET scores for EuroParl and CoVoST2. For each block, best zero-shot scores in \underline{underline} and best in-domain task in \textbf{bold}. Standard deviation presented in \color{gray}{gray}.}
\label{tab:stnew}
\end{table}

\subsection{Spoken Question Answering}\label{experiments:sqa}

\begin{table}
\resizebox{\columnwidth}{!}{\begin{tabular}{llccc}\toprule
&& \textbf{\begin{tabular}[c]{@{}c@{}}Spoken\\ SQuAD\end{tabular}} & \textbf{\begin{tabular}[c]{@{}c@{}}LibriSQA \\ PartI\end{tabular}} & \textbf{\begin{tabular}[c]{@{}c@{}}LibriSQA \\ PartII\end{tabular}} \\\midrule
& Transcripts + EuroLLM 9B \footnotesize{(topline)} & 91.1\%{\footnotesize\color{gray}$\pm2.5$} & 87.6\%{\footnotesize\color{gray}$\pm5.1$} & 73.4\%{\footnotesize\color{gray}$\pm3.1$} \\
& Transcripts + Llama 3.1 8B \footnotesize{(topline)} & 89.2\%{\footnotesize\color{gray}$\pm2.4$} & 85.1\%{\footnotesize\color{gray}$\pm4.5$} & 74.9\%{\footnotesize\color{gray}$\pm3.5$} \\\midrule
& Seamless ASR + EuroLLM 9B \footnotesize{(zero-shot)} & \underline{89.2}\%{\footnotesize\color{gray}$\pm2.9$} & 79.8\%{\footnotesize\color{gray}$\pm6.5$} & 73.5\%{\footnotesize\color{gray}$\pm3.9$} \\
& Seamless ASR + Llama 3.1 8B \footnotesize{(zero-shot)} & 85.6\%{\footnotesize\color{gray}$\pm3.4$} & \underline{82.3}\%{\footnotesize\color{gray}$\pm5.7$} & \underline{74.7}\%{\footnotesize\color{gray}$\pm4.9$} \\
& BEST-IWSLT25-IF \footnotesize{(in-domain)} & 87.4\%{\footnotesize\color{gray}$\pm3.2$} & 80.7\%{\footnotesize\color{gray}$\pm6.7$} & 62.5\%{\footnotesize\color{gray}$\pm4.0$} \\\midrule
\multirow{3}{*}{\rotatebox[origin=c]{90}{\footnotesize{EuroLLM}}}   
& Stage 1 \footnotesize{(zero-shot)} & 61.9\%{\footnotesize\color{gray}$\pm7.4$} & 51.9\%{\footnotesize\color{gray}$\pm15.6$} & 60.3\%{\footnotesize\color{gray}$\pm6.5$} \\
& Stage 2 {[}ASR CE+MSE{]} \footnotesize{(zero-shot)} & \underline{75.1}\%{\footnotesize\color{gray}$\pm9.5$} & \underline{79.3}\%{\footnotesize\color{gray}$\pm6.3$} & \underline{64.3}\%{\footnotesize\color{gray}$\pm4.8$} \\
& Stage 2 {[}ASR/SQA CE{]} \footnotesize{(in-domain)} & \textbf{87.4}\%{\footnotesize\color{gray}$\pm3.2$} & \textbf{83.2}\%{\footnotesize\color{gray}$\pm5.1$} & \textbf{68.1}\%{\footnotesize\color{gray}$\pm2.3$} \\
\arrayrulecolor{gray}\cdashline{2-5}\arrayrulecolor{black}
\multirow{3}{*}{\rotatebox[origin=c]{90}{\footnotesize{Llama 3.1}}} 
& Stage 1 \footnotesize{(zero-shot)} & 62.3\%{\footnotesize\color{gray}$\pm5.1$} & 70.7\%{\footnotesize\color{gray}$\pm7.1$} & \underline{70.5}\%{\footnotesize\color{gray}$\pm3.7$} \\
& Stage 2 {[}ASR CE+MSE{]} \footnotesize{(zero-shot)} & \underline{72.3}\%{\footnotesize\color{gray}$\pm7.6$} & 75.6\%{\footnotesize\color{gray}$\pm7.1$} & 68.9\%{\footnotesize\color{gray}$\pm2.5$} \\
& Stage 2 {[}ASR/SQA CE{]} \footnotesize{(in-domain)} & \textbf{87.9}\%{\footnotesize\color{gray}$\pm3.5$} & \textbf{81.6}\%{\footnotesize\color{gray}$\pm6.0$} & \textbf{72.5}\%{\footnotesize\color{gray}$\pm1.4$} \\\bottomrule
\end{tabular}}
\caption{LLM-as-judge average accuracy for toplines~(top), baselines~(middle) and SpeechMapper models~(bottom). For each block, best zero-shot scores in \underline{underline} and best in-domain task in \textbf{bold}. Standard deviation presented in \color{gray}{gray}.}
\label{tab:new:sqa}
\end{table}

Table~\ref{tab:new:sqa} presents our SQA average accuracy scores computed using LLM-as-judge. We present results for toplines~(Transcript+LLM), pipelines~(ASR+LLM), and BEST-IWSLT25-IF. Due to space constraints, we omit stage-2 CE-only zero-shot IT rows for SpeechMapper, as we observe the same trend from the ST experiments. For in-domain adaptation, and given the short 1K steps training, we stabilized training by sampling ASR data with a probability of 50\%.

We observe that the pipeline systems perform close to the toplines, likely because they leverage the full Seamless model trained on millions of hours. In contrast, BEST-IWSLT-IF and SpeechMapper are end-to-end projectors, trained on much less data, producing \textit{embeddings} that are directly injected into the first LLM layer. Such embeddings are more versatile but may introduce more noise into the LLM backbone than imperfect transcriptions, as LLMs are not exposed to embedding noise during pretraining.

As in the previous section, we observe that stage~1 SpeechMapper models are already capable of performing the target task to a meaningful degree. This highlights that the pretraining stage, which can be run on cheaper hardware, produces a strong, reusable block that only requires limited adaptation, making the subsequent IT stage short and efficient.
For stage-2 SpeechMapper models, and as in the previous section, we again observe that EuroLLM serves as a stronger zero-shot backbone than Llama. Our best zero-shot model rivals BEST-IWSLT25-IF's performance across the LibriSQA test sets, even outperforming it on PartII. This suggests that this baseline, which is trained on SpokenSQuAD, might be generalizing less effectively to datasets with different question styles. 
Moreover, by performing in-domain IT, SpeechMapper not only surpasses this specialist model across all test sets, but also reaches the same performance level as the strong pipeline models. This demonstrates the effectiveness of our task-agnostic pretrained block: with only 1K adaptation steps, it can be quickly specialized to target tasks, delivering strong task-specific performance.
\section{Conclusion}

We introduced SpeechMapper, a two-stage approach for equipping text-only LLMs with zero-shot or task-specific speech-to-text capabilities. SpeechMapper achieves strong zero-shot ST and SQA results, often surpassing strong specialist models, while task-specific IT further improves in-domain performance with minimal data requirements. These results place SpeechMapper as a cost-efficient speech-to-LLM embedding projector solution for speech LLMs.

{
    \small
    \bibliographystyle{ieeenat_fullname}
    \bibliography{main}
}

\clearpage
\appendix
\clearpage
\appendix
\section{Hyper-parameters and Training}
\label{sec:appendix}

\subsection{\speechmapper{} Architecture}\label{app:architecture}
\paragraph{Overall Architecture} \speechmapper{} receives the averaged speech features from a SFM and passes them through the first block containing a CNN layer, a Transformer encoder stack and a fully connected layer. By the end of the first pass of the input through this block, each vector has a dimension of 2048, while the sequence is reduced by half. We then pass the input through a second identical block that reduces the sequence by a half again. This results in output embeddings having 4096-dimensional vectors~(LLM target dimensionality). Lastly, the sequence passes through a 4096x4096 FC layer yielding the final output.

\paragraph{CNN Compression} As mentioned before, \speechmapper{} has a CNN layer in each of its two blocks. Throughout the paper we present results using a model trained with a kernel of 6 and a stride of 2. We experimented reducing this compression by dividing both kernel and stride by two, resulting in a longer output sequence. Then, we adjusted the $\alpha$ parameter to reflect this setting where a larger proportion of the output corresponds to pads. Nonetheless, we did not find a clear advantage in terms of stage~1 ASR scores, and training for stage~1 and 2 for this version is slower due to the longer sequences.

\paragraph{Transformer Stack} Each of our two Transformer encoder stacks consist of 6 attention layers. These encoder stacks have the same architecture as the encoder modules of the original Transformer architecture. We experimented having a more compact stack, with only 3 attention layers, instead of 6. We observed that models with 6 layers systematically outperformed their smaller versions. We believe this hints that capacity is required for learning this complex mapping, since the LLM feedback is not leveraged during this stage, and the SFM used as input is frozen.

\paragraph{Fully Connected Layer} After the input length has been compressed by our CNN layer and it has been passed through the transformer stack, we send the sequence through FC layers to increase the dimensions of the embeddings. This step helps ensure a gradual increase in the vector dimensionality which stabilizes the training. The FC layer in the first block increases the dimensions to 2048 while the FC layer in the second block increases output dimensions to 4096.

\subsection{\speechmapper{} Stage~1 Hyper-parameters} 
\label{section:apdx_sm_stage1}

In this section we discuss the important hyper-parameters from stage~1. Ablation is performed using the ASR task in in-domain~(Librispeech, LS) and out-of-domain~(Voxpopuli, VP) settings. We highlight that some of the tables present \speechmapper{} using non-optimal settings compared the the models presented in the main paper. Unless stated otherwise, we use \llama{} as LLM backbone.

\paragraph{Averaged Features} We experimented with the standard output of \seamless{}, as well as averaging every two or three consecutive frames. Table~\ref{tab:appendix:averageinfo} presents our in- and out-of-domain ASR scores for a \speechmapper{} stage~1 model trained on LS for 1\,M updates with $\alpha=9$.

\paragraph{Speech Model} In this paper we present results using \seamless{} as SFM. Table~\ref{tab:lsonlyresults} presents stage-1 \speechmapper{}s trained using averaged features extracted from the 10th layer of two SSL-based speech representation models: wav2vec~2.0~\cite{NEURIPS2020_92d1e1eb} and mHuBERT-147~\cite{boito2024mhubert}. We observe that SSL models lag behind the model trained using Seamless, especially in out-of-domain settings. We believe this is due to seamless explicitly modeling modality alignment during training~\cite{barrault2023seamlessm4t}.

\paragraph{$\alpha$ Hyper-parameter} We experimented with $\alpha=\{5,7,9\}$. We do not consider removing the padding loss, or having less than half of the MSE weight reserved for the words. Table~\ref{tab:appendix:alphainfo} presents some results for \speechmapper{} trained for 1\,M updates on LS using mHuBERT-147 as SFM. We select $\alpha=5$ mainly due to the results in out-of-domain settings, which we find considerably better than the other options.

\paragraph{$\gamma$ Hyper-parameter} Since the cosine similarity loss ranges between [-1, 1], we observed that setting $\gamma$ to 100 effectively scales the MSE loss closer to our target value. Although our objective is to minimize MSE loss into the single-digit range, the MSE typically remains significantly larger throughout the majority of training, thereby overshadowing the contribution of the cosine loss. Additionally, considering the hyperparameter $\alpha$, which further amplifies the MSE loss by a factor of 10, we selected a $\gamma$ value of 100 consistently across all experiments.

\paragraph{Other Hyper-parameters} We experimented adding dropout of 0.1 during training, which we found to be beneficial. Throughout the paper we present results only for models trained with learning rate of $1e-5$ with a warm-up of 100K steps and an initial learning rate of $1e-8$. We experimented increasing it, which always led to training instabilities, and decreasing it, which led to suboptimal results. For optimizing memory, we use a maximum vector sequence of 1,024 as batch size.

\begin{table}[]
\centering
\resizebox{\columnwidth}{!}{
\begin{tabular}{ccccc}\toprule
\textbf{SFM}         & \textbf{LLM} & \textbf{LS clean}        & \textbf{LS other}        & \textbf{VP}              \\\midrule
Seamless ASR &    --    &     3.4 / 1.2                     &     6.8/ 2.8                     &    7.7 / 5.1                      \\\midrule
mHuBERT-147       & Llama 3.1        & \multicolumn{1}{c}{21.0 / 15.3} & \multicolumn{1}{c}{25.3 / 18.2} & \multicolumn{1}{c}{56.3 / 42.3} \\
wav2vec 2.0& Llama 3.1 & \multicolumn{1}{c}{15.5 / 11.2}  & \multicolumn{1}{c}{17.9 / 11.9} & \multicolumn{1}{c}{49.4 / 37.6} \\
Seamless Encoder & Llama 3.1     & \textbf{9.4} / \textbf{6.5}  & \textbf{12.0} / \textbf{7.9}  & \textbf{25.0} / \textbf{19.7} \\\bottomrule
\end{tabular}}
\caption{ASR WER/CER~($\downarrow$) scores for in- and out-of-domain settings, for stage-1 \speechmapper{} variants in zero-shot settings. 
}
\label{tab:lsonlyresults}
\end{table}
\begin{table}[]
\centering
\resizebox{\columnwidth}{!}{
\begin{tabular}{lcccccc}\toprule
                & \multicolumn{2}{c}{\textbf{LS clean}} & \multicolumn{2}{c}{\textbf{LS other}} & \multicolumn{2}{c}{\textbf{VP}} \\
                & \textbf{WER}      & \textbf{CER}      & \textbf{WER}      & \textbf{CER}      & \textbf{WER}   & \textbf{CER}   \\\midrule
average 3       & 16.3              & 11.5              & 19.9              & 12.8              & 42.3           & 33.7           \\
average 2       & \textbf{13.8}              & \textbf{8.8}               & \textbf{20.7}              & \textbf{13.3}              & \textbf{35.7}           & \textbf{28.8}           \\
standard length & 19.7              & 15.1              & 28.9              & 20.8              & 66.2           & 58.1        \\\bottomrule  
\end{tabular}}
\caption{ASR WER/CER~($\downarrow$) scores for in-domain and out-of-domain settings for stage~1 \speechmapper{}s trained on LS-only with different levels of speech embeddings input compression.}
\label{tab:appendix:averageinfo}
\end{table}
\begin{table}
\centering
\resizebox{\columnwidth}{!}{
\begin{tabular}{lcccccc}\toprule
                & \multicolumn{2}{c}{\textbf{LS clean}} & \multicolumn{2}{c}{\textbf{LS other}} & \multicolumn{2}{c}{\textbf{VP}} \\
                & \textbf{WER}      & \textbf{CER}      & \textbf{WER}      & \textbf{CER}      & \textbf{WER}   & \textbf{CER}   \\\midrule
$\alpha = 5$ & \textbf{23.0}              & 16.0              & 31.3              & 20.3              & \textbf{62.8}          & \textbf{48.3}           \\
$\alpha = 7$ & 23.2              & \textbf{15.7}              & \textbf{30.8}              & \textbf{20.0}              & 66.5           & 56.1           \\
$\alpha = 9$ & 41.4              & 31.0              & 56.3              & 40.8              & 151.2          & 122.5         \\\bottomrule  
\end{tabular}}
\caption{ASR WER/CER~($\downarrow$) scores for in-domain and out-of-domain settings for stage~1 \speechmapper{}s trained on LS with different $\alpha$ loss weights.}
\label{tab:appendix:alphainfo}
\end{table}

\begin{algorithm*}[t!]
\caption{Embeddings Noise Injection}
\label{alg:get-perturbed-embeddings}
\KwIn{%
  \texttt{sentence}\quad\tcp*{string to encode}\\
  \texttt{embeddingLayer}\quad\tcp*{mapping from tokenID to embedding vector}\\
  \texttt{tokenizer}\quad\tcp*{text→tokenID encoder}\\
  \texttt{degreeOfNoise}\quad\tcp*{scale of perturbations; we use 0, e-1, e-2, e-3 and e-4}\\
}
\BlankLine
\KwOut{Stack of perturbed embeddings}
\BlankLine
\BlankLine
promptIDs $\leftarrow$ tokenizer.encode(sentence)\;
noiseList $\leftarrow \{\,i\times \text{degreeOfNoise} \mid i=1,\dots,9\}$\ \quad\tcp*{e.g. [1e-1, 2e-1, ... 9e-1]}
Embeddings $\leftarrow []$\;
\BlankLine
\BlankLine
\ForEach{\text{tokenID} \textbf{in} \textnormal{promptIDs}}{
  emb $\leftarrow \text{embeddingLayer}[tokenID]$\;
  \For{$dimension \gets 1$ \KwTo \textnormal{length[emb]}}{
    $r\gets \text{UniformRandomInt}(1,9)$\;
    emb[$dimension$]$\gets$ emb[$dimension$] + noiseList[$r$]\;
  }
  Append $emb$ to Embeddings\;
}
\BlankLine
\BlankLine
\Return Stack(Embeddings)\;
\end{algorithm*}

\subsection{Stage 1 Training: Embeddings Noise Injection}\label{sec:apdx_noise}

\begin{table}[]
\centering
\resizebox{\columnwidth}{!}{
\begin{tabular}{c|ccccc}
\toprule
LLM & no noise      & $1e-1$      & $1e-2$     & $1e-3$  & $1e-4$  \\ \midrule
Llama 3.1 8B                                & 1.4   & 207.2   & 180.0  & 1.6     & 1.4     \\
EuroLLM 9B                                  & 3.9       &  239.3        & 13.4        & 3.8     & 3.8     \\ \bottomrule
\end{tabular}
}
\caption{LS test clean transcripts WER~($\downarrow$) by adding noise to text embeddings at different precision ranges.}
\label{table:llm_noise}
\end{table}

To learn LLM-like embeddings via MSE, it is essential to understand the target LLM's tolerance to input noise, enabling us to set appropriate training $L_{\text{MSE}}$ targets. To this end, we conduct the following experiment to define what we call the Embedding Error Threshold (EET) for the LLMs used in our study, \llama{} and \eurollm{}. 
First we extracted textual embeddings for the transcriptions from the LS test clean split. We then introduced controlled random noise into their input embeddings, modifying each word vector across all embeddings' dimensions with a given degree of precision. 

Algorithm~\ref{alg:get-perturbed-embeddings} describes the noise injection experiment. For a given noise \textit{degree} $d$, we defined a set of candidate perturbations $\{\,i \times d \mid i \in \{1,\ldots,9\}\}$. This set gave us noise candidates belonging to a particular degree (e.g. $[1e-2, 2e-2, 3e-2,.., 9e-2]$ for $d=e-2$). For each token embedding, we then selected one perturbation at random from this set for a particular degree and added it to every embedding dimension. By varying 
$d$ across orders of magnitude, we determined the decimal precision at which the model's outputs remain stable, i.e., the maximum noise magnitude that does not substantially degrade decoding quality.

We passed these altered sentence embeddings through the LLMs, instructing them to repeat the input text. WER was computed between the generated output and the original transcript. Table~\ref{table:llm_noise} presents the LLMs EET, showing the highest level of embedding noise tolerated without significantly affecting the output.
Both models tolerate noise up to approximately $10^{-3}$, below which they reliably recover the correct text. 

Therefore, if our model produces embeddings matching the target within a value of $10^{-3}$ in each dimension, these should be correctly interpreted by the LLM. 
Then, considering that EETs are a function of distance between the output and target embeddings, the MSE loss becomes a function of the square of EETs. Since our estimated EETs for both the models are $10^{-3}$, our MSE loss target becomes $(\mathrm{EET})^2\approx 10^{-6}$. 
To improve numerical stability, we scale all embedding values by $10^{3}$ before computing the MSE loss. This shifts the MSE loss target to single-digit magnitudes, making the optimization process more stable and efficient.

\subsection{\speechmapper{} Stage~2 Hyper-parameters}\label{appendix:sec:stage2hyperparameters}

\begin{table}
\centering
\resizebox{\columnwidth}{!}{\begin{tabular}{ccccccc}\toprule
                & \multicolumn{2}{c}{\textbf{LS clean}} & \multicolumn{2}{c}{\textbf{LS other}} & \multicolumn{2}{c}{\textbf{VP}} \\
                & \textbf{WER}      & \textbf{CER}      & \textbf{WER}      & \textbf{CER}      & \textbf{WER}   & \textbf{CER}   \\\midrule
$\sigma$ = 0 (CE only) & 4.0               & 1.8               & 7.3               & 3.6               & 11.6           & 6.7            \\
$\sigma$ = 0.2         & \textbf{3.9}               & \textbf{1.6}               & \textbf{6.9}               & \textbf{3.1}               & 12.0           & 6.6            \\
$\sigma$ = 0.4         & 6.2               & 3.5               & 9.6               & 5.5               & 12.5           & 7.6            \\
$\sigma$ = 0.5         & 6.6               & 3.8               & 9.6               & 5.2               & 16.1           & 10.5           \\
$\sigma$ = 0.6         & 4.0               & 1.8               & 7.0               & 3.3               & \textbf{11.1}           & \textbf{6.3}            \\
$\sigma$ = 0.8         & 7.2               & 4.5               & 9.4               & 5.3               & 20.7           & 14.4           \\
$\sigma$ = 0.9         & 4.2               & 1.9               & 7.4               & 3.5               & 12.2           & 7.2            \\
$\sigma$ = 0.95        & 4.8               & 2.2               & 8.2               & 4.0               & 13.1           & 7.4            \\
$\sigma$ = 1 (MSE only)           & 61.5              & 51.4              & 71.5              & 59.4              & 55.7           & 43.7    \\\bottomrule      
\end{tabular}}
\caption{ASR WER/CER~($\downarrow$) scores for in-domain and out-of-domain settings for stage~2 \speechmapper{}s trained on LS-only with different $\sigma$ loss weights.}
\label{tab:stage2:hyperparameters}
\end{table}
\begin{figure*}
  \centering
  \includegraphics[width=1\linewidth]{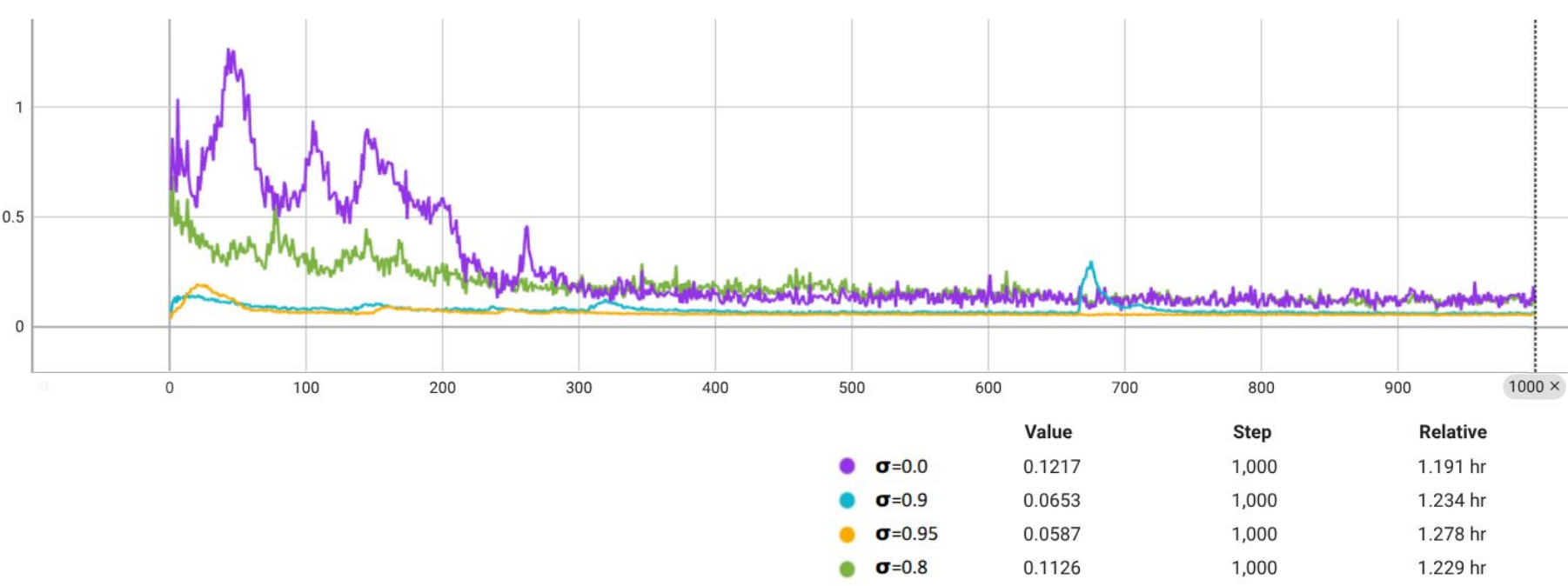}
  \caption{Training curves for stage~2 CE+MSE \speechmapper{}s for four $\sigma$ values.}
  \label{fig:tensorboards}
\end{figure*}

\subsubsection{MSE Loss Weight}
Our IT adaptation step diverges from standard practices by assigning a high weight to $\sigma$, which controls the contribution of the MSE loss. In this section, we justify our choice of $\sigma = 0.9$, which is used for all stage~2 CE+MSE models presented in this paper.

Table~\ref{tab:stage2:hyperparameters} presents scores for stage~2 \speechmapper{} trained with different $\sigma$ values. Some of the training curves are presented in Figure~\ref{fig:tensorboards}. We observe that overall, all $\sigma$ values smaller than one lead to acceptable ASR performance, with $\sigma=1$ adaptation diverging, high values of MSE~($\sigma>0.6$) sometimes presenting small instabilities~(i.e. $\sigma=0.8$), and smaller values resulting in the best WER/CER scores~($\sigma=0.2$). 

To select the appropriate $\sigma$ value, we hypothesized that assigning a higher weight to the MSE loss would reduce overfitting to the IT task~(ASR). 
To assess whether a model is overfitted to a specific task or prompt, we can evaluate it on out-of-domain tasks and instructions. Models that are overfit to the prompt tend to produce ASR outputs regardless of the instruction, whereas prompt-agnostic models adapt and generate the appropriate output for the given task. 

We decoded the first 50 entries of the SpokenSQuAD SQA test set using all models with $\sigma > 0.5$, and manually inspected the outputs to determine whether they corresponded to the expected answers~(SQA) or defaulted to transcripts~(ASR). We observed that models with $\sigma < 0.8$ were unable to produce valid SQA answers, defaulting instead to ASR-like outputs. At $\sigma = 0.8$, the model still showed a high degree of confusion between the SQA and ASR tasks. Only models with $\sigma > 0.8$ consistently produced the expected SQA answers, indicating successful task generalization. 
Therefore, we selected $\sigma=0.9$, which corresponds to the best ASR scores~(see Table~\ref{tab:stage2:hyperparameters}) for a model capable of generalizing to different tasks~($\sigma>0.8$).

\clearpage

\section{Evaluation and Prompts}

\subsection{Other Results}\label{appendix:otherresults}

\paragraph{ASR results} 
Table~\ref{appendix:asrscores} present stage~1 and~2 \speechmapper{} ASR results over four datasets: LS, VoxPopuli~(VP, \citet{voxpopuli}), CommonVoice v17~(CV, \citet{commonvoice:17}), and Fleurs~\cite{conneau2023fleurs}. We observe that EuroLLM stage~1 scores are considerably higher than Llama scores, despite both converging in competitive stage~2 models. After qualitative assessment, we believe EuroLLM's main limitation lies in its verbosity. Despite clear prompt instructions favoring short and direct responses, the model frequently generated multi-line outputs, leading to significant penalties under ASR metrics. In Appendix Section~\ref{appendix:eurollmdecodings} we present generation examples, highlighting in Table~\ref{tab:eurollm:generation} this LLM's tendency to ignore the prompt. In Appendix Section~\ref{sec:werproblem} we discuss limitations of the WER metric in this particular setting.

\paragraph{Language identification}
Table~\ref{app:tab:languageconfusion} present language identification percentage for CE and CE+MSE zero-shot models using their outputs for the ST task from Section~\ref{experiments:st}. We observe that \eurollm{} present less task overfitting, in both CE and CE+MSE settings, compared to \llama{}. We hypothesize this could be a consequence of the backbone itself, as in stage~1, EuroLLM models present considerable less adherence to prompt instructions~(see ASR results discussion above).

\begin{table}
\resizebox{\columnwidth}{!}{\begin{tabular}{llccccc}\toprule
&& \textbf{LS clean} & \textbf{LS other} & \textbf{VP} & \textbf{CV} & \textbf{Fleurs} \\\midrule
& Seamless ASR & \textbf{2.7} / \textbf{0.9}	& \textbf{5.1} / \textbf{2.0} &	\textbf{8.9} / 6.2	& \textbf{8.8} / \textbf{4.1}	& 8.9 / 5.3\\
& BEST-IWSLT25-IF & 4.2 / 1.7	& 7.2 / 3.1 & 	10.3 / \textbf{6.0}	&  9.8 / 4.6	& \textbf{7.4} / \textbf{3.7} \\\midrule
\multirow{3}{*}{\rotatebox[origin=c]{90}{\footnotesize{EuroLLM}}} & stage 1          & 21.5 / 17.5       & 24.0 / 18.5       & 33.9 / 27.4 & 49.5 / 39.9 & 45.3 / 37.5     \\
& stage 2 (ASR CE) & 2.9 / \textbf{1.1}         & 6.0 / \textbf{2.7}         & 11.9 / 7.0  & 15.0 / 8.7  & 11.2 / 7.1      \\
 & stage 2 (CE+MSE) & \textbf{3.1} / 1.2         & \textbf{5.8} / \textbf{2.7}         & \textbf{11.0} / \textbf{6.4}  & \textbf{14.1} / \textbf{8.2}  & \textbf{10.7} / \textbf{6.8}  \\\midrule
\multirow{3}{*}{\rotatebox[origin=c]{90}{\footnotesize{Llama 3.1}}}& stage 1          & 9.4 / 6.5         & 12.0 / 7.9        & 25.0 / 19.7 & 29.2 / 21.3 & 30.2 / 27.6     \\
 & stage 2 (ASR CE) & \textbf{2.9} / \textbf{1.2}         & \textbf{5.8} / \textbf{2.7}         & \textbf{11.8} / \textbf{7.0}  & \textbf{14.6} / \textbf{8.6}  & \textbf{11.9} / \textbf{7.7}      \\
& stage 2 (CE+MSE) & 4.9 / 2.7         & 7.8 / 4.1         & 14.8 / 9.2  & 19.7 / 12.6 & 16.6 / 11.1     \\
\bottomrule
\end{tabular}}
\caption{ASR WER/CER scores for the baseline models~(top) and \speechmapper{} models presented in the main paper. Best scores of each block in \textbf{bold}.}
\label{appendix:asrscores}
\end{table}
\begin{table}[h]
\resizebox{\columnwidth}{!}{\begin{tabular}{lcccc}\toprule
& \multicolumn{2}{c}{\textbf{CE only}} & \multicolumn{2}{c}{\textbf{CE+MSE}} \\
& \textbf{EN \%}     & \textbf{TARGET \%}    & \textbf{EN \%}    & \textbf{TARGET \%}    \\\midrule
\multicolumn{5}{c}{\textbf{\llama{}}}\\
\textbf{EuroParlST en-es} & 31.9\%& 67.1\%& 0.9\% & 98.3\%\\
\textbf{EuroParlST en-fr} & 17.4\%& 82.3\%& 0.3\% & 99.2\%\\
\textbf{EuroParlST en-de} & 38.8\%& 60.5\%& 0.5\% & 99.2\%\\
\textbf{EuroParlST en-it} & 60.8\%& 38.7\%& 0.7\% & 98.9\%\\
\textbf{CoVoST en-de} & 26.1\%& 68.8\%& 4.4\% & 90.3\%\\
\textbf{CoVoST en-zh} & 35.2\%& 42.5\%& 0.8\% & 80.1\% \\\midrule
\multicolumn{5}{c}{\textbf{\eurollm{}}}\\
\textbf{EuroParlST en-es} & 2.2\% & 96.9\%& 1.1\% & 97.9\%\\
\textbf{EuroParlST en-fr} & 1.9\% & 97.8\%& 0.5\% & 99.1\%\\
\textbf{EuroParlST en-de} & 8.8\% & 90.2\%& 2.7\% & 96.2\%\\
\textbf{EuroParlST en-it} & 8.3\% & 90.8\%& 1.1\% & 98.1\%\\
\textbf{CoVoST en-de} & 7.9\%& 85.5\% & 3.3\% & 90.1\%\\
\textbf{CoVoST en-zh} & 5.7\%& 74.5\% & 3.9\% & 76.4\% \\
\bottomrule
\end{tabular}}
\caption{Language identification average percentages by \texttt{langdetect} for all ST test sets, for English and target languages, for both zero-shot adaptation approaches: CE only and CE+MSE. We highlight that for CoVoST, some degree of English is expected, as there is a lot of code-switching in the test set~(e.g. internet terms and brand names).}

\label{app:tab:languageconfusion}
\end{table}

\subsection{On the limitations of WER/CER for LLM-based ASR evaluation}\label{sec:werproblem}

In this work, due to lack of a better metric, and as in similar bibliography in Speech LLMs~\cite{wav2Prompt,hu2024wavllm,salmonn}, we adopt WER/CER as metrics for assessing the capability of our final pipeline to produce transcriptions from corresponding speech. However, we find that this metric overly penalizes our decodings, compared to SFM models such as Whisper or Seamless. In this section we elaborate the reasons, supported by the examples presented in Table~\ref{tab:appendix:werproblem}.

\paragraph{(1) The format complying problem.} In our pipeline, we do not modify the LLM, only producing input embeddings that we request the LLM to repeat~(See prompts at Section~\ref{sec:prompts}). In this setting, the LLM is not forced to comply to a given specific output format, and it might opt towards abbreviations or different forms of expressing numbers. Because of this, correct examples might have their scores penalized.

\paragraph{(2) The strong LLM bias to auto-complete.} We observed during experimentation that in some cases the LLM becomes too biased by the generated tokens it is currently producing, to such an extend that it might ignore the prompt instructions. We observe this tend to happen with book passages, or when the target sentence presents grammatical errors, which the LLM refuses to repeat. In our experiments, we observed that this is a phenomenon that is not limited to our approach: replacing \speechmapper{} embeddings by the reference transcriptions is not enough to push the LLM out of this behavior.

\paragraph{(3) Over-penalization due to the decoding getting stuck.} When running inference on a multimodal prompt, that is, a prompt made of textual tokens and multimodal embeddings, there exist a failure case where the generator of the LLM gets stuck on a given token, repeating it until reaching the limit of tokens set for the inference. This results in a very high mismatch between the target and generated sequence lengths, resulting in WER/CER scores that can reach the thousands. Therefore, a couple of \textit{stuck decodings} might considerably inflates the WER/CER over a given test set.

\subsection{Prompts}\label{sec:prompts}

\paragraph{Embeddings Noise Injection Prompt} 
This is the prompt used for the experiments in Appendix Section~\ref{sec:apdx_noise}. The prompt used for testing the capability of the LLM to decipher the tokens from the noise was: ``\textit{Repeat only the following sentence one more time and nothing else: }''. However, we noticed that the LLMs could be extremely verbose on their answers, impacting WER scores. Therefore, we add additional constraining by adding some tokens after the assistant header to force the LLM to comply with the direct format: ``\textbackslash{}n\textbackslash{}n\textit{Here is the sentence : }''. 

\paragraph{Zero-shot ASR Prompt} Table~\ref{tab:asrprompt} presents our zero-shot ASR prompt. We optimized this prompt by running inference with the reference transcripts of LS test clean on \llama{}, minimizing the WER for the reference text. The reported prompt results in a WER of 0.1, showcasing that even with the reference text, LLMs can fail to comply when asked to repeat input. Further investigation led us to discover that our score different from 0 was due to 9 generated sentences scoring a WER of over 200. In those cases, the LLM was correcting the input text, or completing the content in transcripts~(e.g. biblical passages). 

\paragraph{IT and Zero-shot Evaluation Prompts (ASR/ST/SQA)} We adopt the same prompts from BEST-IWSLT-IF. These prompts were designed to be compact and similar, formulating the instruction in the form of a question. For ST, the instruction is given on the target language to increase the chances of language compliance. Table~\ref{tab:itprompts} presents the template. For each task, the user prompt is assembled by concatenating the prefix with the task prompt and the suffix. These prompts are used for both the stage 2 IT adaptation, and for evaluation. For zero-shot evaluation~(baselines and CE+MSE models), the zero-shot suffix is used to constrain further the LLM generation.

\begin{table*}[]
\centering
\resizebox{\linewidth}{!}{\begin{tabular}{cccc}
\toprule
\multicolumn{1}{c}{\textbf{Target Text}}   & \multicolumn{1}{c}{\textbf{Generated Output}}                                                                        & \textbf{WER}               & \textbf{CER}               \\\midrule
\fix{i am} from the cutter lying off the coast   & \fix{i'm} from the cutter lying off the coast                                                             & 22.2                       & 5.0                        \\\midrule
paul an apostle not of men \fix{et cetera}       & \begin{tabular}[c]{@{}l@{}}paul an apostle not of men \fix{nor through man but through jesus christ and god} \\ \fix{the father who raised him from the dead}\end{tabular}     & 212.5                      & 222.2                      \\\midrule
her sister \fix{is ntombizenhlanhla amanda zuma} & \begin{tabular}[c]{@{}l@{}}her sister \fix{her sister her sister her sister her sister her sister her sister her sister} \\ \fix{her sister her sister her sister her sister her sister her sister her sister her sister} \\ \fix{her sister her sister her sister her sister her sister her sister her sister her sister} \\ \fix{her sister her sister her sister her sister her sister her sister her sister her sister} \\ \fix{her sister her sister her sister her sister her sister her sister her sister her sister} \\ \fix{her sister her sister}\end{tabular} & \multicolumn{1}{r}{1366.7} & \multicolumn{1}{r}{1047.6}\\\bottomrule
\end{tabular}}
\caption{Examples for paragraphs 1-3 from Appendix Section~\ref{sec:werproblem}, with incorrect chunks presented in \fix{red}. Best seen in color.}
\label{tab:appendix:werproblem}
\end{table*}
\begin{table*}
\resizebox{\textwidth}{!}{
\begin{tabular}{l}\toprule
``\textbf{{[}SpeechMapper Embeddings{]}}''\textbackslash{}n\\
Repeat the previous text between the quotes in its entirety just once and nothing else. Do not repeat the text multiple \\times or correct the text or add punctuation. End the text if you notice a phrase or a text is getting repeated. Ignore \\the words that do not make any sense.\\\bottomrule
\end{tabular}}
\caption{Our zero-shot ASR prompt.}
\label{tab:asrprompt}
\end{table*}
\begin{table*}[]
\centering
\resizebox{\textwidth}{!}{\includegraphics[width=1\linewidth]{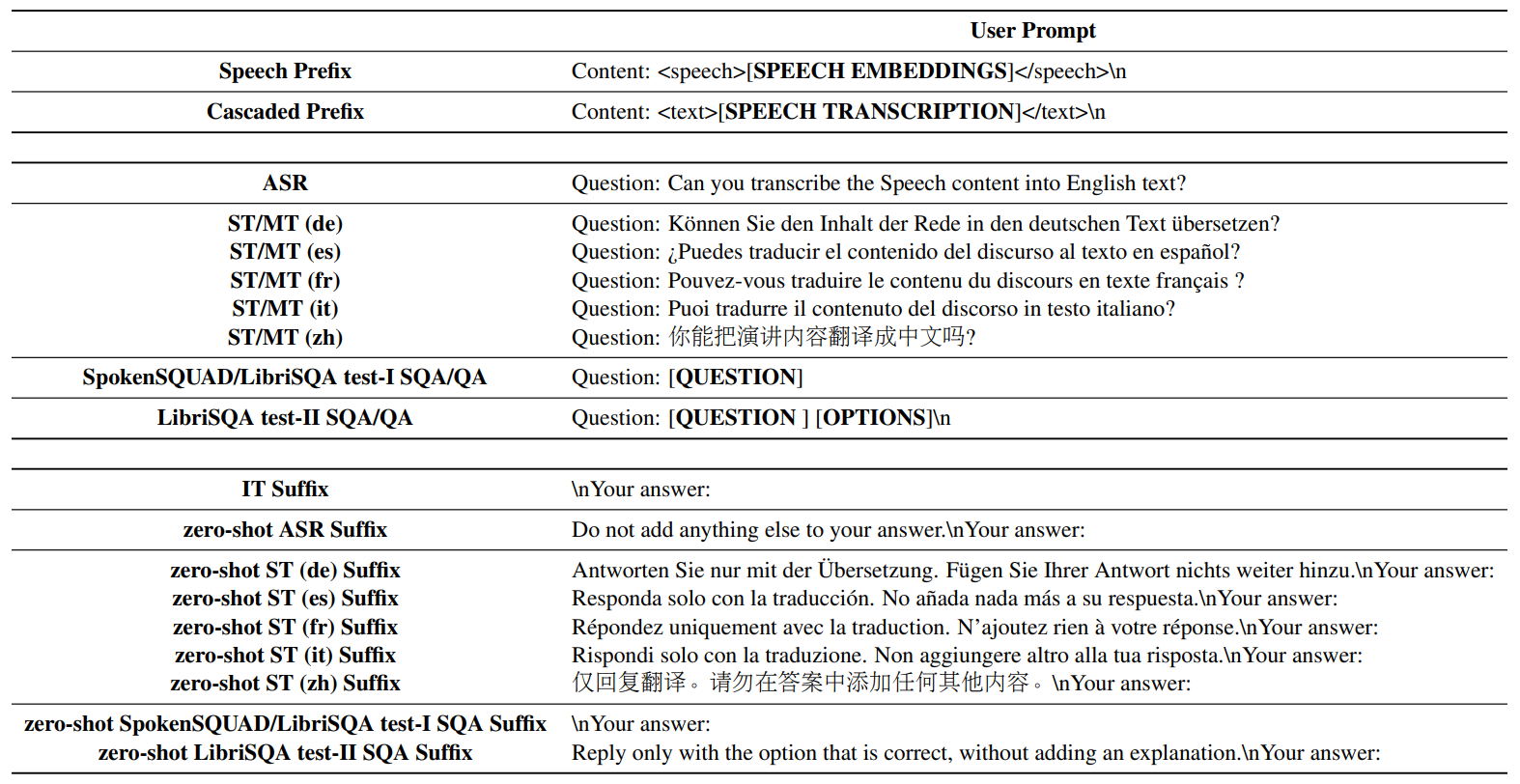}}
\caption{The user turn prompt template used for IT and evaluation. For speech tasks, the user prompt is given by \textbf{Speech Prefix+Task+Suffix}, for cascaded evaluation, the user prompt is given by \textbf{Text Prefix+Task+Suffix}. For zero-shot tasks, the user prompt is given by \textbf{Speech Prefix+Task+zero-shot Suffix}. \\ST/MT instructions were obtained by translating the instruction ``\textit{Can you translate the Speech content into \textbf{{[}German/Spanish/French/Italian/Chinese{]}} text?}'' to corresponding target languages using \seamless{}.}
\label{tab:itprompts}
\end{table*}

\clearpage

\section{Generation Examples}

In this section we provide decoding examples for the embeddings generated by \speechmapper{} Stage~1 models.

\subsection{EuroLLM decodings}\label{appendix:eurollmdecodings}

We observe that EuroLLM is a particularly challenging model to control during generation. It often answers in unexpected format, when compared to Llama 3.1. Table~\ref{tab:eurollm:generation} presents some examples. In general, we observe that in many cases the LLM refuses to give the expected answer, while in other occasions, the provided answer is correct, but produced around many other elements that were not requested. We hypothesize this happens for two reasons. Firstly, this LLM seems to be less tolerant to noisy embeddings~(from \speechmapper{}) compared to Llama 3.1. Secondly, the instructed model seems to be aligned in such a way that the model can more easily deviate from instructions. We performed several experiments by varying system and user prompts, but we were unable to find a setting where EuroLLM would behave in a similar manner to Llama 3.1.

\begin{table*}
\resizebox{\linewidth}{!}{
\begin{tabular}{c|c|c}\toprule
\textbf{Issue}& \textbf{Generation}& \textbf{Target}\\\midrule
\begin{tabular}[c]{@{}c@{}}Repeating the prompt \\ instead of the text\end{tabular}     & \begin{tabular}[c]{@{}c@{}}i will repeat the previous text between the quotes in its \\ entirety just once and nothing else do not repeat the text \\ multiple times or correct the text or add punctuation\end{tabular} & \begin{tabular}[c]{@{}c@{}}scanner manufacturers usually have their \\ own name attached to this technique\end{tabular}           \\ \midrule
\multirow{3}{*}{\begin{tabular}[c]{@{}c@{}}Refusal to comply \\ to prompt\end{tabular}} & \begin{tabular}[c]{@{}c@{}}the text was too long to repeat in its entirety so i'm going \\ to leave it as is\end{tabular}& the dental clicks are also laminal denti alveolar\\\cline{2-3}
& i rephrased the text in its entirety& then leave with the caravan\\\cline{2-3}
& \begin{tabular}[c]{@{}c@{}}the sentence has several issues but one of the main ones is its unclear \\ context the original text appears to be a fragment of a story or \\ a statement but there's not enough information to understand \\ what it's referring to however\end{tabular}          & \begin{tabular}[c]{@{}c@{}}after the second sino japanese war broke out \\ wu refused to cooperate with the japanese\end{tabular} \\\midrule
\begin{tabular}[c]{@{}c@{}}Including \\ explanations\end{tabular}& \begin{tabular}[c]{@{}c@{}}tred out leslie herself dropped asleep\textbackslash{}n\textbackslash{}n please note \\ that the text was just repeated once as per your request\end{tabular}& tired out leslie herself dropped asleep\\\midrule
\begin{tabular}[c]{@{}c@{}}Incorrect word \\ separator\end{tabular}& \begin{tabular}[c]{@{}c@{}}alexander went\textbackslash{}n\textbackslash{}nover\textbackslash{}n\textbackslash{}nand\textbackslash{}n\textbackslash{}nopened\textbackslash{}n\textbackslash{}nthe\\ \textbackslash{}n\textbackslash{}nwindow\textbackslash{}n\textbackslash{}nfor\end{tabular} & \begin{tabular}[c]{@{}c@{}}alexander went over and opened the \\ window for her\end{tabular}\\\bottomrule
\end{tabular}
}
\caption{Example of generations produced by EuroLLM using \speechmapper{} embeddings.}
\label{tab:eurollm:generation}
\end{table*} 

\subsection{Common Mistakes}
In this section we present and discuss some of the common mistakes produced by prompting LLMs using \speechmapper{} embeddings. Table~\ref{tab:llms:generation} presents these examples, while the next paragraphs discuss them in more details.

\begin{table*}
\resizebox{\linewidth}{!}{
\begin{tabular}{c|c|c}\toprule
\textbf{Issue}& \textbf{Generation}& \textbf{Target}\\\midrule
\multirow{3}{*}{\begin{tabular}[c]{@{}c@{}}Synonym \\ replacement\end{tabular}} & galleries later sold part of his \textbf{inheritance} & the gallery later sold parts of his \textbf{legacy} \\\cline{2-3}
& \begin{tabular}[c]{@{}c@{}}the novel's \textbf{quite} straightforward narrative \\ belies the attitudes of the main characters\end{tabular} & \begin{tabular}[c]{@{}c@{}}the novel's \textbf{fairly} straightforward narrative \\ belies the attitudes of the main characters\end{tabular} \\\cline{2-3}
& \begin{tabular}[c]{@{}c@{}}if you want to be close to the action you're \\ going to have to get in early to get a camping \\ \textbf{spot} close to the music\end{tabular} & \begin{tabular}[c]{@{}c@{}}if you want to be close to the action you're \\ going to have to get in early to get a camping \\ \textbf{site} close to the music\end{tabular} \\\midrule
\multirow{3}{*}{\begin{tabular}[c]{@{}c@{}}Issues with \\ named entities\end{tabular}} & he met peter \textbf{flfl} during his time there & he met peter \textbf{fluck} during his time there \\\cline{2-3}
& \begin{tabular}[c]{@{}c@{}}subsequently obtained permission \\ to return to ireland\end{tabular} & \begin{tabular}[c]{@{}c@{}}\textbf{maccarthy} subsequently obtained permission \\ to return to ireland\end{tabular} \\ \cline{2-3}
& the parish includes the village of \textbf{sivton montes} & the parish includes the village of \textbf{sutton montis} \\\cline{2-3}
& we're not using wagons \textbf{murork} told him line them up & we're not using wagons \textbf{murdoch} told him line them up \\\midrule
\multirow{2}{*}{\begin{tabular}[c]{@{}c@{}}Pronoun \\ replacement\end{tabular}}        & \textbf{i} will attend columbia university & \textbf{banerjee} will attend columbia university \\
& this was \textbf{my} plan & this was \textbf{your} plan \\\midrule
\multicolumn{1}{c|}{\multirow{2}{*}{Repetitions}} & he \textbf{also also} time in paris and london & he \textbf{also spent} time in paris and london \\
 & i don't believe all i hear no not by \textbf{bigbig} deal & i don't believe all i hear no not by \textbf{a big} deal\\\bottomrule
\end{tabular}}
\caption{Example of generations produced by LLMs using \speechmapper{} embeddings.}
\label{tab:llms:generation}
\end{table*}

\paragraph{Synonym Replacement} We believe synonym replacement happens during generation due to \speechmapper{} learning in the embedding space. Since semantically similar words are often close in the embedding space, the produced embeddings can be \textit{ambiguously in between} the embeddings for some words. We highlight this mistake might often not be an issue: if the exact word is not needed, the model will still be able to perform ST or SLU tasks from these embeddings, as the meaning is often maintained.

\paragraph{Named Entities} One of the main limitations of our approach is regarding named entities. Since we train directly on the embedding space, \speechmapper{} is unable to correctly transcribe entities it has never seen during training. We observe that due to this, the model will either try to piece together the word using sub-word units, or produce very noisy embeddings that tend to be ignored by our zero-shot ASR generation due to the prompt instructions. In the future we intend to explore manners to mitigate this limitation.

\paragraph{Pronoun/Subject Change} Surprisingly, we observed that prompting an LLM to repeat/copy text~(ASR task) between quotes can be challenging. Sometimes the LLMs produce the target text by modifying the pronoun, as it was itself \textit{repeating} the text. We investigated several prompting variations, but we were unable to completely mitigate this issue.

\paragraph{Repetitions} \speechmapper{} pre-training forces the semantic information inside a sequence to be moved to the first vectors, filling the rest of the sequence with the pad embedding. We observe that during the decoding of less-performing models, the generations present repetitions of words or sub-words. In many cases, a given word appears twice, while the following word is omitted. We believe this is a consequence of the alignment-free learning of embeddings: two consecutive vectors might encode the same word. In future work we intend to investigate using an ASR loss to reduce repetition.

\clearpage
\section{Licenses}\label{apdx_license}

\subsection{SFM Backbones}
\begin{itemize}
    \item \seamless{} \hfill \texttt{CC-BY-NC-4.0}
    \item \texttt{mHuBERT-147} \hfill \texttt{CC-BY-NC-SA-4.0}
    \item \texttt{Wav2Vec 2.0} \hfill \texttt{Apache-2.0}
\end{itemize}

\subsection{LLM Backbones}
\begin{itemize}
    \item \llama{} \hfill \texttt{Meta Llama 3.1 License}
    \item \eurollm{} \hfill \texttt{Apache-2.0}
\end{itemize}

\subsection{Training Dataset}
\begin{itemize}
    \item \textbf{LibriSpeech} \hfill \texttt{CC-BY-4.0}
\end{itemize}

\subsection{Libraries}
\begin{itemize}
    \item \texttt{Pasero} \hfill \texttt{CC-BY-NC-SA-4.0}
    \item \texttt{Torchtune} \hfill \texttt{BSD 3-Clause}
\end{itemize}

\section{Pre-trained Modules}\label{sec:pretrained}

Table~\ref{tab:modelsurl} lists all the pretrained models used in this work, for either training~(top) or evaluation~(bottom), with links for download.  We do not use any close model in this work.

\begin{table}[ht]
\resizebox{\columnwidth}{!}{
\begin{tabular}{ll}
\toprule
\textbf{Model}                      & \textbf{URL}                                                                \\\midrule
\texttt{mHuBERT-147}          & utter-project/mHuBERT-147\\
\texttt{wav2vec2-base-960h}        & facebook/wav2vec2-base-960h\\
\seamless{} & facebook/seamless-m4t-v2-large\\
\texttt{Llama-3.1-8B-Instruct } & meta-llama/Llama-3.1-8B-Instruct \\

\texttt{EuroLLM-9B-Instruct} & utter-project/EuroLLM-9B-Instruct \\\midrule
\texttt{wmt22-comet-da} & Unbabel/wmt22-comet-da
\\
\texttt{Llama-3.1-70B-Instruct} & meta-llama/Llama-3.1-70B-Instruct \\
\texttt{Gemma3-12B-Instruct} & google/gemma-3-12b-it \\
\texttt{Gemma3-27B-Instruct} & google/gemma-3-27b-it \\
\bottomrule
\end{tabular}}
\caption{Pre-traned modules and HuggingFace IDs.}
\label{tab:modelsurl}
\end{table}

\clearpage

\end{document}